\documentclass{article}

\usepackage[utf8]{inputenc} 
\usepackage[T1]{fontenc}    
\usepackage{arxiv}
\usepackage{hyperref}       
\usepackage{url}            
\usepackage{booktabs}       
\usepackage{amsfonts}       
\usepackage{nicefrac}       
\usepackage{microtype}      
\usepackage{lipsum}
\usepackage{graphicx}
\usepackage{multicol}
\usepackage{xcolor}
\usepackage{ulem}

\title{LiSSS: A toy corpus of Spanish Literary Sentences for  Emotions detection}

\author{
  Juan-Manuel Torres-Moreno\\
  Université d'Avignon/LIA\\Polytechnique Montréal\\
  \texttt{juan-manuel.torres@univ-avignon.fr}
  \And 
  Luis-Gil Moreno-Jiménez\\
  Université d'Avignon/LIA\\Universidad Tecnol\'ogica de la Selva\\
  \texttt{luis-gil.moreno-jimenez}\\
  \texttt{@alumni.univ-avignon.fr}
}

\begin{document} 

\maketitle

\begin{abstract}
In this work we present a new small data-set in Computational Creativity (CC) field, the Spanish Literary Sentences for emotions detection corpus  (LISSS). 
    We address this corpus of literary sentences in order to evaluate or design  algorithms of emotions classification and detection. 
    We have constitute this corpus by manually classifying the sentences in a set of emotions: Love, Fear, Happiness, Anger and Sadness/Pain. 
    We also present some baseline classification algorithms applied on our corpus.
    The LISSS corpus will be available to the community as a free resource to evaluate or create CC-like algorithms.
\end{abstract}
     

\section{Introduction}

Research in Natural Language Processing (NLP) focused in the classification of emotions,
has used corpus constituted by encyclopedic documents (mainly Wikipedia), journals (newspapers or magazines) or specialized
(legal, scientific or technical documents) for the development and evaluation of the models.
\cite{torres2014,cunhaCSSMV11,sierra}.
Studies of literary corpus have been systematically left aside mainly because the level of literary discourse is more complex than other genres.
In this work we introduce a new literary emotion corpus in order to evaluate and validate NLP algorithms in  the literary emotions classification tasks.

This paper is structured as follows.
In Section \ref{sec:arte} we show some works related to development and analysis of corpora.
In Section \ref{sec:corpus} we describe the corpus LISSS and in Section \ref{sec:modelos+results} the learning corpus \textit{CiteIn}.
Four \textit{baseline} classification models are described in Section \ref{sec:results}, as well as their respective results.
Finally in Section \ref{sec:conclusiones}, we propose some ideas for future works before to conclude.

\section{Related work}
\label{sec:arte} 

Several corpora in Spanish have been built and made available to the scientific community \cite{acllaw:2011}
however, a few number of them have been classified considering categories of emotions.
For example the corpus SAB, composed by tweets in Spanish was introduced in \cite{corpusSAB}.
The tweets represent critics toward different commercial brands. The annotation was made considering the emotion perceived for each tweet. 
The corpus SAB consists of 4~548 annotated tweets using 8 predefined emotions: [\textit{Trust, Satisfaction, Happiness, Love, Fear, Disaffection, Sadness and Anger}].

Another data set concerning tweets is the corpus TASS \cite{corpusTASS}.
It contains about 70~000 tweets classified using automatic methods into the following categories:
[\textit{Positive, Negative, Neutral, None}]. 
Tweets of the TASS corpus are related with different topics: Politics, Economy, Sport, Music, etc.).
In \cite{chen2014building} it is presented a global analysis (at word level) about emotions polarity. 
The corpus employed is composed of several lexicons in 40 languages, including Spanish.
Annotation considered the categories: [\textit{Positive and Negative}] with a score in the range between 0 and 1 (0 defines a word as negative and 1 as positive).

\section{The LiSSS corpus}
\label{sec:corpus}

Unlike the above-mentioned corpora, we introduced in this paper a Literary Sentiment Sentences in Spanish corpus named LISSS.
It consists only of literary texts, which gives it a particular characteristic more useful for studying the algorithms of automatic classification and generation of literary text.
Moreover, for the classification, five categories of emotions were defined, instead of a binary (positive-negative) classification. 
This characteristic of LISSS could be useful for more complete analysis.

\subsection{Corpus structure}
\label{constitucion:LISSS}

Our work was meticulous but on a modest scale.
We decided to create a small controlled corpus, exclusively composed of literary phrases in Spanish selected from universal literature.
The LISSS corpus was constituted manually using literary texts in Spanish coming from 140 Spanish-speaking authors or using authors translations from  languages other than Spanish (the full list of authors studied is available in the Annex).
Each phrase in the corpus was read and manually classified using the following five categories:
\begin{itemize}
	\item Anger (A)
	\item Love (L)
	\item Fear (F)
	\item Happiness (H)
	\item Sadness or Pain (S)
\end{itemize}

Since the sentences (or group of sentences) could belong to two or more emotions, we included the possibility of labelling the sentences using all the emotions expressed.
The corpus currently has 230 sentences. 
Sentences were selected manually to maintain a balance between the five types of emotions.
Sentences were processed to create a text document coded in utf8\footnote{Versions XML and POS tagged using Freeling 4.1 (\url{http://nlp.lsi.upc.edu/freeling/node/1}) are also available.}.
Each line of the corpus contains the information of the following fields:

\begin{itemize}
	\item ID
	\item Sentence (or paragraph)
	\item \# Author
\end{itemize}

Each field is separated by a tab character.
The ID field is composed of a sequential number (1,2,3,...) followed by a code (A, L, F, H, S) for each of the predefined emotions.
If s sentence is emotionally ambiguous, it will have as many codes as categories it belongs to.

It should be noted that sentences are actually sometimes mini-paragraphs composed of several sentences. This was done to respect as much as possible the coherence of the idea expressed and the corresponding emotion. For example, sentence 24 of the emotion Love (L), by Lope de Vega:

\begin{verbatim}
80L  La raíz de todas las pasiones es el amor. De él nace la tristeza, el gozo, 
     la alegría y la desesperación.       # Lope de Vega
\end{verbatim}

is a two-sentence paragraph:

\begin{verbatim}
     La raíz de todas las pasiones es el amor. 

     (The root of all passions is love.)

     De él nace la tristeza, el gozo, la alegría y la desesperación. 

     (From it they are born sadness, joy, happiness and despair.)
     
\end{verbatim}
 
\vfill 
 
\subsection{Characterization of LISSS corpus}
\label{caracteristicas:LISSS}

The corpus contains 230 literary sentences (or paragraphs in a line).
It was constituted manually from selected quotes, stories, novels and some poems.

The literary genre is homogeneous, but several topics are present in the corpus.
Sentences in general language as well as those too short ($N \le 3$ words) or too long ($N \ge 50$ words) were carefully avoided.
Finally we got a vocabulary complex and aesthetic where certain literary figures like anaphora or metaphor could be observed.
The characteristics of LISSS corpus are showned in Table~\ref{tab:1}.
\begin{table}[htb]
  \centering
  \begin{tabular}{lc|rrrr}
    \toprule
   &\bf Label & \bf Sentences 	&\bf Paragraphs &\bf Words& \bf Characters\\
    \midrule
   \bf Corpus&\bf LISSS& 230&26		& 4 577	& 25 643 \\ \hline
    Anger	  &A	& 43	&7 		& 961 	& 5 428 \\
    Love 	  &L    & 43	&8 		& 941 	& 5 166 \\ 
    Fear	  &F	& 46	&5 		& 940 	& 5 173 \\     
    Happiness &H	& 46	&2 		& 889 	& 5 143 \\     
    Sadness/Pain&S	& 41	&7 		& 846 	& 4 728 \\             
    \bottomrule
  \end{tabular}
  \caption{Corpus LISSS of literary sentences classified in 5 emotions.}
  \label{tab:1}
\end{table}

10\% of the LISSS corpus is constituted by mini-paragraphs composed of several sentences.
The five existing classes are not completely homogeneous given the existence of ambiguous sentences that belong to two or more emotions Table \ref{tab:2}).

\begin{table}[htb]
  \centering
  \begin{tabular}{lc|c}
    \toprule
     \bf Emotion  & \bf Label & \bf Number of multi-emotional sentences \\
    \midrule
    Anger	  &A	& 8\\
    Love 	  &L    & 12 \\ 
    Fear	  &F	& 7 \\     
    Happiness &H	& 4 \\     
    Sadness/Pain&S	& 5 \\  \hline
    TOTAL	  &	& 36 \\    
    \bottomrule
  \end{tabular}
  \caption{Corpus LISSS: Multi-emotional sentences.}
  \label{tab:2}
\end{table}

An example is the sentence labeled with the identifier 14AL, Anger (A) and Love (L):
\begin{verbatim}
14AL	Del amor al odio, solo hay mas amor	# Mario Benedetti

(From love to anger, there's only more love)
\end{verbatim}

that belongs to both categories.

In total, 36 sentences ($\approx$ 16\%) of LISSS corpus are considered with more than one emotion.
This ambiguity is mainly observed in the emotions Love and Sadness.
Literary ambiguity represents a challenge to automatic classification methods.

The LISSS corpus has the advantage of being homogeneous in terms of genre by possessing only sentences considered as ``literary sentences'', but it is heterogeneous in terms of emotions classes.

In others emotion corpora, the sentences may be in general language: sentences that often give a fluency to the reading and provide the necessary relations between ideas expressed in literary sentences.
Likewise, the corpora with tweets are not able to be used with literary goals due to the presence of \textit{noise} and other special characters.
Another advantage of LISSS corpus is that the presence of noise (cut phrases, pasted words, wrong syntax, etc.) was avoided by a repeated and carefully reading.

However, LISSS corpus has the disadvantage of having a reduced size, this is not suitable to algorithms that use automatic learning.
This is normal, because the goal of LISSS corpus is not to be used for learning, but for evaluation
The LISSS corpus is suitable for testing the quality and performance of such algorithms.

The LISSS corpus version 0.230 (06/06/2020) is available on website: \url{http://juanmanuel.torres.free.fr/corpus/lisss/} under GPL3 public licence.

\section{Classification algorithms employed}
\label{sec:modelos+results}

In this section we  present four classical classification methods applied on the LISSS corpus and some preliminary results.

\subsection{Employed models}

The LISSS corpus was tested with several classical classification algorithms available in the Weka's system framework and libraries\footnote{\url{https://www.cs.waikato.ac.nz/ml/weka/}}. In particular, we have employed:

\begin{itemize} 
	\item Algorithm J48 (decision trees of C45 type) 

	\item Naive Bayes, classical implementation

	\item Naive Bayes Multinomial (NBM), implementation oriented to textual classification
	
	\item Support Vector Machine (SVM)

\end{itemize}

The J48 algorithm, proposed by Ross Quinlan \cite{salzberg1994c4}, belongs to the family of models based on decision trees.
It is an extension of the ID3 algorithm.
Its analysis is based on the search for information entropy and was considered for its high performance in classification tasks.

We also decided to use the Naive Bayes model given its wide implementation in several classification processes. 
In particular, we used the Naive Bayes Multinomial model, based on the calculation of estimated frequency of terms, which allows a simple and efficient implementation in textual classification \cite{su2011large}.

Finally, we tested with a standard implementation of a SVM\footnote{\url{https://weka.sourceforge.io/doc.stable/weka/classifiers/functions/LibSVM.html}} to compare the performance of models on the LISSS corpus.

The four algorithms used need a learning phase to produce a classification model. 
The learning phase must be done on a corpus independent of the test corpus. 
In our case we decided to build a learning corpus suitable for this task, adapting it to the five categories present in the LISSS corpus.

\subsection{Learning corpus}

For the training of classification models, we built an ad hoc learning corpus.
The \textit{CitasIn} corpus composed of texts in Spanish, mostly from the literary genre. 
A large number of documents belonging to different categories\footnote{\url{https://citas.in/temas/}} (friendship, lovers, beauty, success, happiness, laughter, enmity, deception, anger, fear, etc.) were retrieved from a suitable website\footnote{All documents were downloaded, with the editor authorisation, on 25 March 2020 from the website: \url{https://citas.in}}.
These documents were classified in the five classes of the LISSS corpus, from a manual mapping with their own categories (last column of the Table \ref{tab:3}).

The generated corpus has an adequate size to be used in training tasks of automatic learning models.
The disadvantage of the corpus \textit{CitasIn} is the
presence of noise given that it often contains supporting sentences (sentences with general language vocabulary) that do not belong to the literary genre.
The characteristics of the corpus\textit{CitasIn} are found in the Table~\ref{tab:3}.  The reader should have no problem reconstituting the corpus \textit{CitasIn} using this correspondence between class.

\begin{table}[htb]
  \centering
  \begin{tabular}{l|cccc|c}
    \toprule
 \bf CitasIn  & \bf Sentences  &\bf Words	& \bf Chars& \bf Words per   &   \bf Categories  \\
   &    &               &               &\bf  sentence &\url{https://citas.in/temas/}\\ \midrule
   \bf Corpus & 53 351  & 1 903 214 & 11 113 527 & 35.6 & \\ \hline
    Love	  & 13 430	& 392 623   & 2 234 523  & 29.2 & \it alma, amantes, amistad, amor\\
              &         &           &     &             & \it belleza, beso, esperanza, pasion \\ \hline
    Happiness & 10 857  & 377 280 	& 2 222 645 & 34.7  & \it felicidad, amistad, diversión, sonrisa, motivación\\
              &         &           &           &      & \it risa, victoria, exito, optimismo \\  \hline
    Anger     & 9 556 	& 384 211 	& 2 253 546 & 40.2& \it egoismo, enemistad, engaño, envidia, venganza\\
              &     &   &           & & \it guerra, infierno, mentira, odio, muerte \\ \hline   
    Fear	  & 8 917	& 355 976   & 2 093 918 & 39.9 & \it necesidad, miedo, dolor, fracaso\\
              &   &     &           &   & \it indecisión, problema, soledad, suicidio \\ \hline
    Sadness/Pain&10 591 & 393 124 	& 2 308 895 & 37.1 & \it despedida, tristeza, pena, enfermedad, fracaso\\
              & & & &  & \it pérdida, sufrimiento, olvidando, llorar, lágrima \\ \bottomrule
  \end{tabular}
  \caption{Corpus \textit{CitasIn}: Sentences from different categories grouped into 5 emotions}
  \label{tab:3}
\end{table}

We pre-processed the corpus \textit{CitasIn} before the learning phase.
The texts were coded in utf-8 format, we removed the special symbols, as well as the stop words using the Weka libraries and stop lists.
We normalized the words by transforming the capital letters into small letters. 
Finally a tokenization process was applied using Weka specific algorithms for Spanish language.
Of course, we have eliminated from 
the \textit{CitasIn} corpus, the common sentences with the LISSS corpus. 

\section{Baseline results and discussion}
\label{sec:results}

In this section we show the baseline results of our tests. 
The tests were executed with the four models presented in Section \ref{sec:modelos+results}, using \textit{CitasIn}  as learning corpus and LISSS corpus (see Section \ref{sec:corpus}) for evaluation.
In all cases, there were 53 351 input instances (sentences) for training the systems.
Our results also show the confusion matrix calculated for each algorithm.

\subsection{Algorithm Trees J48}

\begin{verbatim}
=== Summary ===

Correctly Classified Instances          146               58.4    %
Incorrectly Classified Instances        104               41.6    %
Total Number of Instances               250     

=== Detailed Accuracy By Class ===

                 Precision  Recall   F-Measure  Class
                 0.478      0.880    0.620      Love
                 0.700      0.700    0.700      Happiness
                 0.615      0.640    0.627      Anger
                 0.500      0.240    0.324      Sadness
                 0.719      0.460    0.561      Fear
Weighted Avg.    0.602      0.584    0.566          

=== Confusion Matrix ===

  a  b  c  d  e   <-- classified as
 44  2  1  2  1 |  a = Love
  6 35  3  2  4 |  b = Happiness
 11  1 32  5  1 |  c = Anger
 22  8  5 12  3 |  d = Sadness
  9  4 11  3 23 |  e = Fear
\end{verbatim}

\subsection{Algorithm Naive Bayes Multinomial Text}

\begin{verbatim}
=== Summary ===

Correctly Classified Instances         128               51.2    %
Incorrectly Classified Instances       122               48.8    %
Total Number of Instances              250     

=== Detailed Accuracy By Class ===

                 Precision  Recall   F-Measure  Class
                 0.338      0.880    0.489      Love
                 0.657      0.460    0.541      Happiness
                 0.629      0.440    0.518      Anger
                 0.737      0.280    0.406      Sadness
                 0.806      0.500    0.617      Fear
Weighted Avg.    0.633      0.512    0.514                  

=== Confusion Matrix ===

  a  b  c  d  e   <-- classified as
 44  4  0  2  0 |  a = Love
 24 23  1  0  2 |  b = Happiness
 24  2 22  0  2 |  c = Anger
 26  4  4 14  2 |  d = Sadness
 12  2  8  3 25 |  e = Fear
\end{verbatim}

\subsection{Algorithm Support Vector Machine}

\begin{verbatim}
=== Summary ===

Correctly Classified Instances         124               49.5      %
Incorrectly Classified Instances       126               50.4      %
Total Number of Instances              250     

=== Detailed Accuracy By Class ===

                 Precision  Recall   F-Measure  Class
                 0.304      0.760    0.434      Love
                 0.714      0.400    0.513      Happiness
                 0.653      0.640    0.646      Anger
                 0.471      0.160    0.239      Sadness
                 0.839      0.520    0.642      Fear
Weighted Avg.    0.596      0.496    0.495  

=== Confusion Matrix ===

  a  b  c  d  e   <-- classified as
 38  2  2  5  3 |  a = Love
 29 20  1  0  0 |  b = Happiness
 14  1 32  1  2 |  c = Anger
 35  4  3  8  0 |  d = Sadness
  9  1 11  3 26 |  e = Fear
\end{verbatim}

\subsection{Algorithm Naive Bayes}

\begin{verbatim}
=== Summary ===

Correctly Classified Instances          91               36.4    %
Incorrectly Classified Instances       159               63.6    %
Total Number of Instances              250     

=== Detailed Accuracy By Class ===

                 Precision  Recall   F-Measure  Class
                 0.238      0.780    0.364      Love
                 0.645      0.400    0.494      Happiness
                 0.550      0.220    0.314      Anger
                 0.444      0.080    0.136      Sadness
                 0.654      0.340    0.447      Fear
Weighted Avg.    0.506      0.364    0.351           



=== Confusion Matrix ===

  a  b  c  d  e   <-- classified as
 39  3  2  1  5 |  a = Love
 25 20  0  1  4 |  b = Happiness
 37  2 11  0  0 |  c = Anger
 41  3  2  4  0 |  d = Sadness
 22  3  5  3 17 |  e = Fear
\end{verbatim}

The best model in this task was the J48 tree algorithm, obtaining an average F-measure = 0.566 (harmonic combination of Precision and Recall). This relatively low result shows the difficulty of the task of classifying emotions in literary corpora.

We detected two main problems in the classification of this type of texts; on the one hand, the richness of the lexicon used. On the other hand, the ambiguity. Belonging to multiple emotions in the same sentence causes the errors of the methods used. In particular the emotions Sadness and Love 
were confused even with the best algorithm, J48.

\section{Conclusion and future work}
\label{sec:conclusiones}

In this article we have introduced a new toy literary corpus of emotions in Spanish, the LISSS corpus.
The aim of this corpus is to test machine learning algorithms on a specialized corpus, no to train such algorithms.
We have tested four classic classification algorithms on our LISSS corpus.
The results obtained show that the classification of this type of text is a difficult exercise.
The sentences often belong to two or more classes.
The overlap between the vocabulary of the different classes, causes the methods to be unable to correctly classify this corpus.

We think that automatic classifiers can be enriched through the integration of other modules, using characteristics of language, style or personality detection to achieve a better classification  \cite{We12, plastino2016fisica, Ed17, Si18, moreno_GEN}.

Future work needs the introduction of more sentences in order to enrich the corpus.
The scientific community can contribute to develop this corpus, modify it or distribute it under the GPL3 license. 

\section*{Acknowledgements}

This work is funded by Consejo Nacional de Ciencia y Tecnología (Conacyt, Mexico), grant number 661101 and partially by the Université d'Avignon/Laboratoire Informatique d'Avignon (LIA), France. 
We thank the admin of the site \url{https://cite.in} for the facilities in allowing us to use their literary quotation database for our experiments.
Also, authors thank Carlos-Emiliano Gonz\'alez-Gallardo for their comments and invaluable corrections of this paper.

\section*{Annex: Author Listing of LISSS corpus version 0.230}

\begin{multicols}{3}
Abraham Lincoln;
Agatha Christie;
Albert Camus;
Albert Einstein;
Albert Schweitzer;
Aldous Huxley;
Alphonse Daudet;
Alphonse de Lamartine;
Alphonse Karr;
Amado Nervo;
Anais Nim;
Anatole France;
Andrés Calamaro;
Antoine de Saint-Exupéry;
Arthur Schopenhaue;
Baruch Spinoza;
Benjamin Franklin;
Bernard Le Bouvier de Fontenelle;
Bertrand Russell;
Blaise Pascal;
Buda;
Camilo José Cela;
Cesare Pavese;
Charles Baudelaire;
Charles Bukowski;
Charles Dickens;
Charles Péguy;
Charly García;
Cleóbulo de Lindos;
Denis Diderot;
Dostoievsky;
Edgar Allan Poe;
Elbert Hubbard;
Epicteto de Frigia;
Ernest Hemingway;
Ernesto Che Guevara;
Ernesto Sábato;
Federico García Lorca;
Fiodor Dostoievski;
F Nietzche;
François de La Rochefoucauld;
Gabriel García Márquez;
George Bernard Shaw;
George Orwell;
George Sand;
George Steiner;
Giacomo Leopardi;
Gilbert Keith Chesterton;
Giordano Bruno;
Goethe;
G Patton;
Graham Greene;
Groucho Marx;
Gustave Flaubert;
Heinrich Heine;
Henry Louis Mencken;
Hermann Hesse;
Honoré de Balzac;
HP Lovecraft;
Immanuel Kant;
Isaac Asimov;
Italo Calvino;
Jacinto Benavente;
Jaime Sabines;
Jane Addams;
Jean Cocteau;
Jean-Jacques Rousseau;
Jean Luc Goddard;
Jean Paul Sartre;
Jim Morrison;
John F Kennedy;
John Lennon;
Jorge Luis Borges;
José Ingenieros;
José Marti;
José Saramago;
Juan Manuel Torres Moreno;
Juan Ramón Jiménez;
Juan Rulfo;
Jules d'Aurevilly;
Laura Esquivel;
Leonardo Da Vinci;
Lope de Vega;
Lord Byron;
Marcel Proust;
Marie Curie;
Mario Benedetti;
Mark Twain;
Marlene Dietrich;
Marqués de Vauvenargues;
Martin Luther King;
Maximilien Robespierre;
Máximo Gorki;
Miguel de Cervantes;
Miguel de Unamuno;
Miguel Hernández;
Milan Kundera;
Molière;
Montesquieu;
Nelson Mandela;
Nietzsche;
Ogden Nash;
Orson Welles;
Oscar Wilde;
Ovidio;
Pablo Neruda;
Paulo Coelho;
Pedro Bonifacio Palacios Almafuerte;
Pericles;
Peter Alexander Ustinov;
Pierre Corneille;
Proverbio chino;
Ray Loriga;
René Descartes;
R Tagore;
Sabino Arana;
Sadamm Hussein;
Selma Lagerlöf;
Séneca;
Shakespeare;
Simone de Beauvoir;
Sir Francis Bacon;
Sófocles;
Solón;
Stanisław Lem;
Stephen Hawking;
Sthendal;
Susan Sontag;
Tennessee Williams;
Terencio;
Tito Livio;
Tupac Shakur;
Ugo Foscolo;
Victor Hugo;
Vinicius de Moraes;
Voltaire;
William Blake;
William Faulkner;
William Nicholson;
Woody Allen
\end{multicols}

\bibliographystyle{plain}
\bibliography{biblio.bib}
\end{document}